\documentclass[10pt,twocolumn,letterpaper]{article}

\usepackage{cvpr}
\usepackage{times}
\usepackage{epsfig}
\usepackage{graphicx}
\usepackage{amsmath}
\usepackage{amssymb}

\usepackage{booktabs}
\usepackage{algorithm}
\usepackage{algorithmic}

\usepackage[sort]{cite}

\usepackage{url}
\usepackage{flushend}
\usepackage{amsmath}
\usepackage{subfigure}
\usepackage[utf8]{inputenc}
\usepackage[english]{babel}

\usepackage[pagebackref=true,breaklinks=true,letterpaper=true,colorlinks,bookmarks=false]{hyperref}

\cvprfinalcopy 

\def\cvprPaperID{5033} 

\ifcvprfinal\fi
\begin{document}

\title{MBS: Macroblock Scaling for CNN Model Reduction}

\author{Yu-Hsun Lin, Chun-Nan Chou, and  Edward Y. Chang\\
HTC Research \& Healthcare (DeepQ)\\
{\tt\small \{lyman\_lin, jason.cn\_chou, edward\_chang\}@htc.com}
}

\maketitle

\begin{abstract}
In this paper we propose the macroblock scaling (MBS) algorithm, which can be applied to various CNN architectures to reduce their model size.  
MBS adaptively reduces each CNN macroblock depending on its information redundancy measured by our proposed  effective flops.
Empirical studies conducted with ImageNet and CIFAR-$10$ attest that MBS can reduce the model size of some already compact CNN models, e.g., MobileNetV$2$ ($25.03\%$ further reduction) and ShuffleNet ($20.74\%$), and even ultra-deep ones such as ResNet-$101$ ($51.67\%$) and ResNet-$1202$ ($72.71\%$) with negligible accuracy degradation.
MBS also performs better reduction at a much lower cost than the state-of-the-art optimization-based methods do.
MBS's simplicity and efficiency, its flexibility to work with any CNN model, and its scalability to work with models of any depth make it an attractive choice for CNN model size reduction.
\end{abstract}
\section{Introduction}

CNN models have been widely used by image-based applications, thanks to the breakthrough performance of AlexNet \cite{alexnet}.  
However, a very deep and wide CNN model consists of many parameters, and as a result, the trained model may demand a large amount of DRAM and a large number of multiplications to perform a prediction.  
Such high resource and computation requirements lead to latency, heat, and power consumption problems, which are suboptimal for edge devices such as mobile phones and IoTs \cite{survey-efficient-dnn}.
Therefore, reducing CNN model size is essential for improving resource utilization and conserving energy.

Several CNN model reduction algorithms have recently been proposed \cite{survey-efficient-dnn}. 
These algorithms can be divided into two categories:  micro-level (performing reduction/quantization inside a filter) and macro-level reduction (removing redundant filters).
These two categories are complement to each other. (More details are presented in the related work section.)  We focus our study in this paper on macro-level reduction. 
For lucid exposition, we establish an essential concept that removing redundant filters are also referred as removing redundant channels since each filter outputs a corresponding channel in a CNN model \cite{mobilenet}.

There are two categories of macro-level reduction: \textit{optimization based} and \textit{channel-scaling based}. 
Each category has multiple methods and algorithms.
The optimization-based category typically estimates the filter importance by formulating an optimization problem with the adopted criteria (e.g., filter weight magnitude). 
Removing a filter (or a channel, which is formally defined in Section~\ref{sec-MBS}) will affect both the former and latter layers.
The filter pruning step of the optimization-based method must take into account the inter-connected structures between CNN layers.
Therefore, a CNN model such as DenseNet \cite{densenet} and ShuffleNet \cite{shufflenet} with more complex inter-connected structures may prevent the optimization-based approach from being effective. 

The channel-scaling based category uses an $\alpha$-scalar to reduce channel width.  
For instance, MobileNet \cite{mobilenet} uses the same $\alpha$-scaler to prune the widths of all channels. 
Applying the same $\alpha$-scaler to all convolutional layers without considering each information density is a coarse-grained method.
A fine-grained method that finds the optimal $\alpha$-scalar for each convolutional layer should be ideal. 
However, the increasingly complicated inter-layer connection structures of CNN models forbid fine-grained scaling to be computationally feasible.  


To address the shortcomings of the current model-compaction methods, we propose \textit{macroblock scaling} (MBS). 
A macroblock consists of a number of convolution layers that exhibit similar characteristics, such as having the same resolution or being a segment of convolution layers with customized inter-connects.
Having macroblock as a structure abstraction provides the flexibility for MBS to inter-operate with virtually any CNN models of various structures, and also permits channel-scaling to be performed in a ``finer''-grained manner.
To quantify information density for each macroblock so as to determine an effective macroblock-dependent scalar, MBS uses {\em effective flops} to measure each macroblock's information density. (We define {\em effective flops} to be the number of convolution flops required for the activated non-zero ReLU outputs.)
Experimental results show that the reduction MBS can achieve is more significant than those achieved by all prior schemes.

Our contributions can be summarized as follows:

\begin{itemize}
	\item  MBS employs {\em macroblock} to address the issues that both coarse-grained and fine-grained scaling cannot deal with, and hence allows channel-scaling to be performed with any CNN models.
    \item MBS proposes using an effective and efficient measure, {\em effective flops}, to quantify information density to decide macroblock-dependent scaling factors. As shown in the algorithm section, the complexity of MBS is linear with respect to the number of training instances times the number of parameters, which is more efficient than the optimization-based methods \cite{nisp,prune-filter}.
    \item Extensive empirical studies on two representative datasets and various well-known CNN models (e.g., MobileNet, ShuffleNet, ResNet, and DenseNet) demonstrate that MBS outperforms all state-of-the-art model-reduction methods in reduction size while preserving the same level of prediction accuracy.
Due to its simple and effective nature, MBS remains to be effective even with ultra-deep CNNs like ResNet-$101$ on ImageNet ($51.67\%$ reduction) and ResNet-$1202$ on CIFAR-$10$ ($72.71\%$ reduction).
\end{itemize}

The remaining parts of this paper are organized into three main sections.
The Related Work section highlights some previous efforts 
of reducing CNN models.
The Method section explains our proposed MBS algorithm.
The Experiment section shows the encouraging results 
of applying MBS on various CNN models.



\section{Related Work}


We review related work in two parts.
We first review key CNN properties relevant to the inception of
MBS and then review some 
representative model-reduction methods.

\subsection{Relevant CNN Properties}


There are research works \cite{branchynet,adaptive-computation,blockdrop} integrating the early stop (or early exit) mechanism of the initial CNN layers in order to speed up the inference process.
This phenomenon demonstrates that the outcome of a CNN model at early stages can be adequate for predicting an image label with high confidence.
This result provides supporting evidence for us
to group convolution layers into two types: former convolution layers (near to the input image) as the base layers, and latter convolution layers (close to the label output) as the enhancement layers. 
The early stop mechanism motivates that the information
density in the enhancement layers should be lower
than that in the base layers, and therefore, 
more opportunities exist in the enhancement layers to reduce model size.

\subsection{Reduction Methods of CNN Models}

As mentioned in the introduction that model reduction can be divided into
micro-level and macro-level approaches. 
Binary approximation of a CNN filter is one important direction for micro-level model reduction \cite{binary-connect,xnor-net,abcnet,binarynet}.
Maintaining prediction accuracy of a binary CNN is a challenging issue \cite{aaai-accurate-binarynet}. 
The sparse convolution modules \cite{sparse-conv,structured-sparse,low-rank,linear-structure,net-trim} or deep compression \cite{deep-compression} usually introduce irregular structures.
However, these micro-level model reduction methods with irregular structures often require special hardware for acceleration \cite{eie}. 

The macro-level model reduction approach removes irrelevant filters and maintains the existing structures of CNNs \cite{prune-filter,iccv-prune-channel,network-slim,second-order-prune}.
The methods of this reduction approach estimate the filter importance by formulating an optimization problem based on some adopted criteria (e.g., the filter weight magnitudes or the filter responses). 

The research work of \cite{nisp} addresses the filter importance issue by formulating the problem into binary integer programming with the aid of feature ranking \cite{inf-fs}, which achieves the state-of-the-art result.
For an $n$-convolution-layer model with $n_p$ parameters and $N$ training images, the complexity of acquiring the CNN outputs is $O(n_p \times N)$. 
The complexity of the feature ranking step is $O(N^{2.37})$ \cite{inf-fs}.
In addition to the preprocessing step, the binary integer programming is an NP-hard problem. 
The detail complexity is not specified in \cite{nisp}. 
In general, a good approximate solution for $n_p$ variables still requires high computational complexity (e.g., $O(n_{p}^{3.5})$ by linear programming \cite{linear-programming}).
In contrast, MBS enjoys low complexity that is $O(n_p \times N)$ in computing information density and $O(M \times n)$ in computing the scaling factors for a CNN with $n$ convolution layers and $M$ macroblocks, which is detailed in the next section.  

In a sense, some model reduction methods belonging to network pruning are related to the topic of architecture search.
Our proposed MBS also falls into this type.
Treating network pruning as architecture search only requires a one-pass training, but the search space is restricted to all sub-networks inside a large network \cite{rethink-pruing}.
By contrast, full architecture search such as DARTS \cite{liu2018darts} considers more options, e.g., activation functions or different layer orders, and usually pays the cost of requiring more passes to find the goal architecture.
Thus, applying architecture search on large datasets like ImageNet directly demands considerable computation cost.

\begin{figure*}[t]
	\begin{center}
		\centerline{\includegraphics[width=0.8\textwidth]{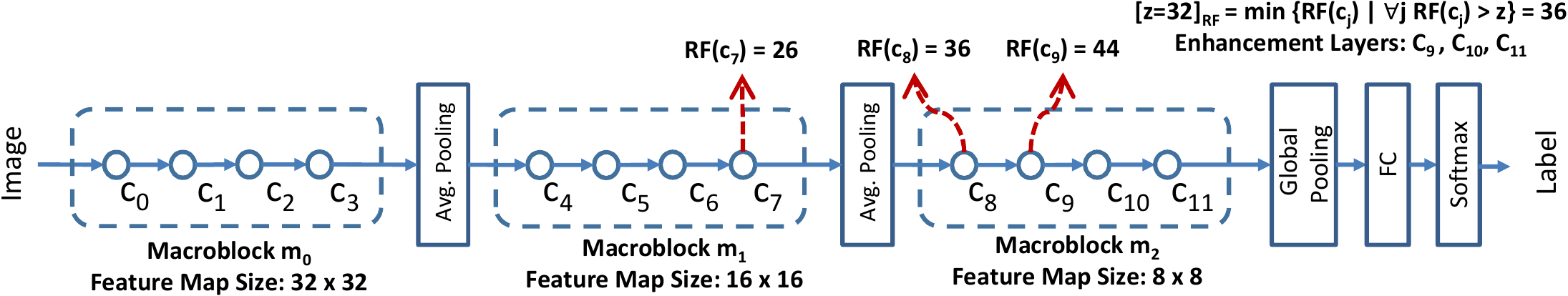}}
		\caption{An example CNN model contains three CNN macroblocks for the CIFAR image ($32 \times 32$ pixels). Each macroblock $m_i$ consists of the convolution layers whose output feature maps are of the same size (i.e., same width and height). Besides, $ \text{k\_size}_{j} = 3 \ \forall j$.}
		\label{fig-model}
	\end{center}
\end{figure*}

\section{MBS Algorithm}
\label{sec-MBS}

This section presents our proposed macroblock scaling (MBS) algorithm for reducing an already trained CNN model.  We first define key terms including {\em channel}, {\em filter}, {\em channel scaling}, {\em macroblock}, and the parameters used by MBS.  We then explain how MBS computes information density, and how that information is used by MBS to reduce model size. 
Finally, we analyze computational complexity of MBS and compare its efficiency with competing model-compaction algorithms.  

Let us use image applications to explain a CNN pipeline.
A typical CNN pipeline accepts $N$ training images as input. These $N$ training instances are of the same height and width. To simplify our notation, we assume all input images are in the square form with the same resolution $L \times L$.  
A CNN model is composed of multiple convolution layers.  
The input to a convolution layer is a set of input tensors (or input activations), each of which is called a {\em channel} \cite{survey-efficient-dnn}. Each layer generates a successively high-level abstraction of the input tensors, call a output tensor or feature map.  
 
More specifically, the $j^{\textup{th}}$ convolution layer $c_j$, $j=0,\dots,n-1$, takes   $\text{s\_size}_{j} \times \text{s\_size}_{j} \times \text{c\_width}_{j}$ input tensor and produces $\text{s\_size}_{j+1} \times \text{s\_size}_{j+1} \times \text{c\_width}_{j+1}$ output tensor, where $\text{s\_size}$ is the spatial size of input or output tensors, and $\text{c\_width}$ is the input/output channel width (i.e., number of channels). 
Particularly, $\text{s\_size}_{0}$ is equal to $L$.
Let $\text{k\_size}_{j}$ denote the spatial size of the square kernel of $c_j$, the required number of parameters of $c_j$ can be written as
\begin{equation} \label{eq-conv-parameter}
\text{k\_size}_{j} \times \text{k\_size}_{j} \times \text{c\_width}_{j} \times \text{c\_width}_{j+1}.
\end{equation}

MBS groups convolution layers into {\em macroblocks}.  
A macroblock consists of the convolution layers whose output tensors (feature maps) are of the same size.
We use $M$ to denote the number of macroblocks in a CNN model.
Figure \ref{fig-model} depicts an example CNN model with $M=3$.
The size of output tensors is down-sampled by the pooling 
layers with stride size 2.
Hence, macroblock $m_i$ is defined as
\begin{equation} \label{eq-macroblock}
m_i = \{  c_j \ | \forall j \  s.t. \ \text{s\_size}_{j+1}=\frac{L}{2^i} \  \}.
\end{equation}

Operation {\em scaling} reduces channel width. 
Intuitively, MBS would like to prune channels that cannot provide positive contributions to accurate prediction.  
For instance, MobileNet \cite{mobilenet} scales down all channel widths by a constant ratio $0 < \alpha < 1$, or we call this baseline scheme $\alpha$-scaling.
MobileNet uses the same $\alpha$ value for all convolution layers.
However, an effective channel-scaling scheme should estimate the best scaling ratio for each convolution layer based on its {\em information density}

But quantifying and determining the scaling ratio for each convolution layer disturbs the designs of CNN models.
To preserve the design structure of an original CNN model, MBS performs reduction at the macroblock level instead of at the convolution-layer level.  
For some CNN models that have  convolution layers connected into a complex structure, MBS treats an entire such segment as a macroblock to preserve its design.  
Our macroblock approach, as its name suggests, does not deal with detailed inter-layer connection structure. 
The macroblock abstraction thus makes model reduction simple and adaptive.

\begin{figure}[t]
 	\begin{center}
 		\centerline{\includegraphics[width=0.75\columnwidth]{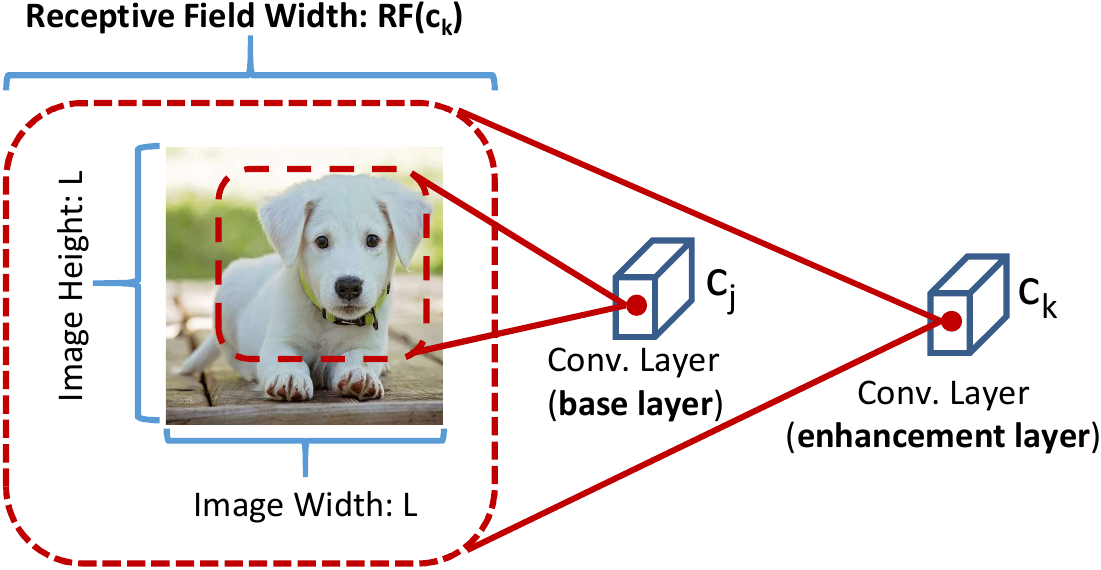}}
 		\caption{An example of the receptive field of the neuron in the base layer $c_j$ and the enhancement layer $c_k$.}
 		\label{fig-field}
 	\end{center}
\end{figure}

\subsection{Grouping Convolution Layers by Receptive Fields}

An effective CNN model requires a sufficient number of convolution layers to capture good representations from input data.
However, as the number of the convolution layers grows beyond a threshold, the additional benefit in improving prediction accuracy can diminish.
One may argue that the former convolution layers may learn low-level representations such as edges and contours, whereas latter layers high-level semantics. 
As we will demonstrate shortly, the latter layers may cover {\em receptive fields} that are larger than the input images, and their learned information may not contribute to class prediction.
The effective receptive field in CNN is the region of the input image that affects a particular neuron of the network \cite{receptive-field}.
Figure~\ref{fig-field} shows an example, where a neuron of a former convolution layer covers a region inside the input image, whereas a neuron of a latter layer may cover a region larger than the input image.
Hence, we categorize convolution layers into two types, base layers and enhancement layers, which are defined as follows:
\begin{itemize}
	\item Base convolution layers:   
    The former convolution layers (near to the input) of a CNN model learn  essential representations from the training data. Though representations captured in the base layers could be redundant, they are fundamental for class prediction.
	
	\item Enhancement convolution layers:
    The latter convolution layers may cover receptive fields larger than the input areas\footnote{Due to data augmentation and boundary patching operations applied to raw input images, a training input image may contain substantial useless information at its boundaries.}. Therefore, opportunities are available for channel pruning to remove redundant information.
\end{itemize}

\begin{figure}[t]
	\begin{center}
		\centerline{\includegraphics[width=0.8\columnwidth]{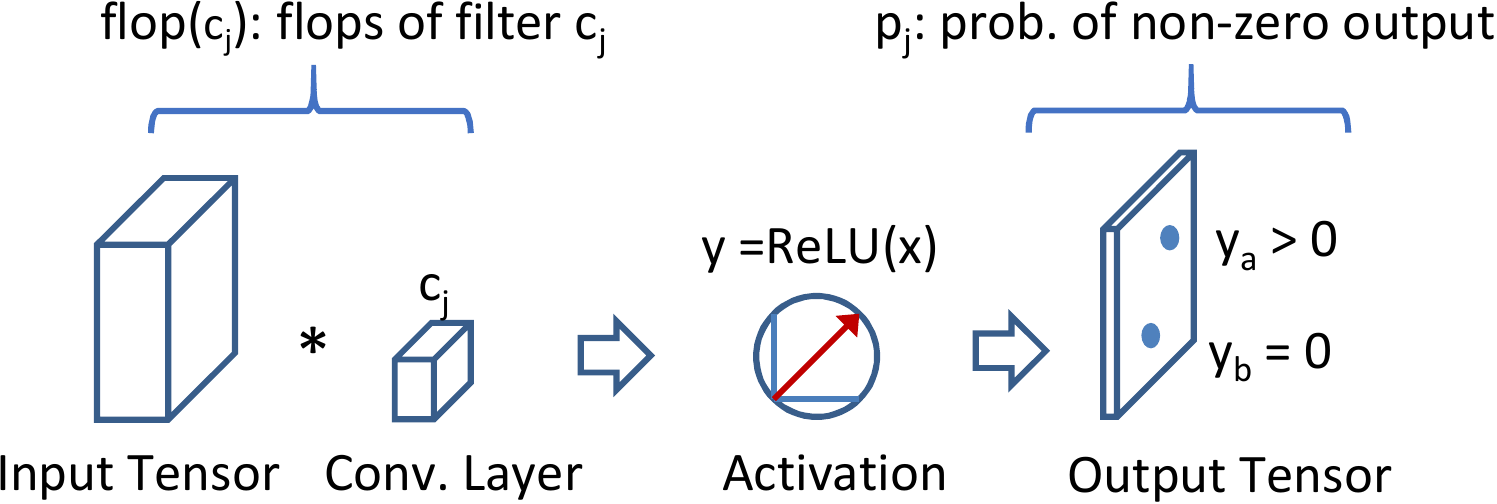}}
		\caption{Effective flop calculates of the flops considering the non-zero probability $p_j$ of the ReLU output.}
		\label{fig-relu}
	\end{center}
\end{figure}

Now, we define a function $\textup{RF}(c_j)$ to compute the receptive field size of layer $c_j$.
For simplicity, assume that the receptive field region of filter $c_j$ is $\textup{RF}(c_j) \times \textup{RF}(c_j)$.
The possible set of values of $\textup{RF}(c_j)$ is discrete, which is determined by the configuration of the kernel size and the stride step of a CNN model.
For lucid exposition, we define $\lceil z \rceil_{\textup{RF}}$ to characterize the minimum receptive field boundary that is greater than a given value $z$ as follows:
\begin{equation}\label{eq-boundary}
\lceil z \rceil_{\textup{RF}} = \min \{\textup{RF}(c_j) \ | \ \forall j \ \textup{RF}(c_j) > z \}.
\end{equation}
We use this boundary $\lceil z \rceil_{\textup{RF}}$ to divide base convolution layers and enhancement convolution layers in a CNN pipeline.

Revisit macroblocks in Figure \ref{fig-model}. 
Convolution layer $c_{9}$ belonging to macroblock $m_2$ is the first enhancement layer, where its receptive field is larger than the receptive field boundary $\lceil z =32 \rceil_{\textup{RF}} = 36$. 
We estimate information redundancy of each macroblock $m_i$ by measuring the information density ratio contributed by the enhancement layers.

We can determine the base layers of a CNN by setting the value of $z$.
As we have previously explained, the area beyond and at the boundary of an image contains less useful information.
Thus, setting $z=L$ is reasonable to separate those layers that can contribute more to class prediction from the other layers.
A macroblock can contain base layers only, enhancement layers only, or a mixture of the two.

\subsection{Information Density Estimation within \\ Macroblocks by Effective Flops}
MBS uses {\em convolution FLOP} to estimate information density.
A FLOP (denoted by the lowercase ``flop'' in the remainder of the paper) is a multiply-and-add operation in convolution.
The more frequently that ReLU outputs a zero value means that the less information that convolution layer contains.
Therefore, only those flops that can produce a non-zero ReLU output are considered to be effective.

A neuron on a convolution layer covers a region, i.e., receptive field of the input image.
The physical meaning of effective flops represents effective feature matchings on the input image whose the feature pattern size is the same as that of the  receptive fields.
Since effective flops quantify the effective feature matchings of CNN models for a given image, it plays the key role in estimating information density.

Figure \ref{fig-relu} shows the computation of the effective flops of a convolution layer.
Let $e_{c_j}$ denote effective flops of layer $c_j$, and $p_j$ represent the non-zero probability of its ReLU output.
We can define $e_{c_j}$ as
\begin{equation}\label{eq-flops}
e_{c_j} = p_j \times \textup{flop}(c_j).
\end{equation}

To evaluate information density of macroblock $m_i$, we tally the total effective flops from the beginning of the CNN pipeline to the end of macroblock $m_i$.
We can write the sum of the effective flops as
\begin{equation}\label{eq-model-flops}
E_{\textup{total}}(m_{i})  = \sum e_{c_j},\ c_j \in \{m_0, \cdots, m_i \}.
\end{equation}
Next, we compute the effective flops in the base layers or those flops taking place within the receptive field as  
\begin{equation}\label{eq-compact-flops}
E_{\textup{base}}(m_{i}) = \sum e_{c_j}, \ c_j \in \{ \textup{RF}(c_j) \leq \lceil z \rceil_{\textup{RF}} \},
\end{equation}
where the base layers have the maximum receptive field size $\lceil z \rceil_{\textup{RF}}$.

Based on the total flops $E_{\textup{total}}(m_{i})$ and base flops $E_{\textup{base}}(m_{i})$, we define the difference between the two as the enhancement flops, which is denoted as $E_{\textup{enchancement}}(m_{i})$ and can be written as $E_{\textup{total}}(m_{i}) - E_{\textup{base}}(m_{i})$.  
The redundancy ratio $r_i$ of macroblock $m_i$ is then defined as the total enhancement flops over the total flops, or 
\begin{equation}\label{eq-r}
r_i = \frac{E_{\textup{enhancement}}(m_{i})}{E_{\textup{total}}(m_{i})} = 1 - \frac{E_{\textup{base}}(m_i)}{E_{\textup{total}}(m_{i})}.
\end{equation}
We estimate the channel-scaling factor for each macroblock $m_i$ based on this derived redundancy $r_i$, which is illustrated in the next subsection.

\begin{algorithm}[t] 
\caption{Macroblock Scaling} 
\label{alg-mb-scaling} 
\begin{algorithmic}
	\REQUIRE  $F_n()$, $I_{0 \sim N-1}$ /*Pre-trained model, training images\\
    \ENSURE $[\text{c\_width}_{m_0}^c,\cdots, \text{c\_width}_{m_{M-1}}^c]$/*Compact model 
     \textbf{Procedure:}
     \begin{itemize}
     \item $\textup{NZ}()$ /*Computes the number of non-zero elements
     \item $\textup{RF}()$ /*Computes receptive field size
     \item $\textup{flop}()$ /*Computes the number of FLOPs
    \end{itemize}
	\textbf{Variable:}
    \begin{itemize}
    \item $v_j^{I}$ /*The $j^{\textup{th}}$ ReLU output tensor of $F_n(I)$
    \end{itemize}
    
\end{algorithmic}
\textbf{BEGIN}
\begin{algorithmic}[1] 
	\FOR{$j=0, \cdots, n-1$} 
    	\FOR{$i=0,\cdots,N-1$}	
        \STATE $p^{I_i}_j \gets \frac{\textup{NZ}(v_j^{I_i})}{\text{s\_size}_{j+1} \times \text{s\_size}_{j+1} \times \text{c\_width}_{j+1} }$
    	\ENDFOR
    \STATE $p_j \gets \frac{1}{N} \sum p^{I_i}_j $ /*Compute non-zero output prob.
    \STATE $e_{c_j} \gets  p_j \times \textup{flop}(c_j) $ /*Effective flops for each $c_j$
    \ENDFOR
	\FOR{$i=0, \cdots, M-1$}
    \STATE $E_{\textup{total}}(m_{i}) \gets 0$ /*Initialization
    \STATE $E_{\textup{base}}(m_{i}) \gets 0$ /*Initialization
    \FOR{$j=0, \cdots, n-1$}
    	\IF{$c_j \in \{m_0, \cdots, m_i \}$}
    	\STATE $E_{\textup{total}}(m_{i})  \gets E_{\textup{total}}(m_{i}) +  e_{c_j}$ /*Total $e_{c_j}$ 
    	\ENDIF
        \IF{$c_j \in \{ \textup{RF}(c_j) \leq \lceil z \rceil_{\textup{RF} } \}$}
    	\STATE $E_{\textup{base}}(m_{i}) \gets E_{\textup{base}}(m_{i}) +  e_{c_j}$ /*Base layer $e_{c_j}$  
    	\ENDIF
\ENDFOR
        \IF{$E_{\textup{total}}(m_{i}) > E_{\textup{base}}(m_{i}) $ }  
        \STATE $r_i \gets 1 - \frac{E_{\textup{base}}(m_i)}{E_{\textup{total}}(m_{i})} $ /*Compute redundancy 
          \STATE $\beta_i \gets \frac{1}{1+r_i}$
        \ELSE{}
          \STATE $\beta_i \gets 1$
        \ENDIF
        \STATE $\text{c\_width}_{m_{i}}^c \gets \lceil \beta_i \times \text{c\_width}_{m_{i}} \rceil$ /*Compact model
    \ENDFOR
    \STATE return $[\text{c\_width}_{m_0}^c,\cdots, \text{c\_width}_{m_{k-1}}^c]$ 
\end{algorithmic}
\textbf{END}
\end{algorithm}

\subsection{Channel-Scaling Factor Estimation}

We define the relation between the original channel width  $\text{c\_width}_{m_i}$ of macroblock $m_i$ and the compact channel width $\text{c\_width}_{m_i}^c$ after the reduction process, which is depicted as 
\begin{equation}\label{eq-w}
\text{c\_width}_{m_i} = (1+r_i) \times \text{c\_width}_{m_i}^c.
\end{equation}
If there is no redundancy in macroblock $m_i$ (i.e., $r_i = 0$), the original channel $\text{c\_width}_{m_i}$ is equal to the compact channel width $\text{c\_width}_{m_i}^c$.
Therefore, the channel width multiplier 
$\beta_i$ for the macroblock $m_i$ is
\begin{equation}\label{eq-beta}
\beta_i = \frac{1}{1+r_i},
\end{equation}
where this estimation makes  $\beta_i > 0.5$ since $r_i <1$ according to Eq. (\ref{eq-r}).
The lower bound of the channel-scaling factor $\beta_i$ is in accordance with the observation made by MobileNet \cite{mobilenet} that a scaling factor that is less than $0.5$ can introduce noticeable distortion.

Algorithm \ref{alg-mb-scaling} presents our MBS algorithm, which estimates the scaling factor $\beta_i$ for each macroblock $m_i$ and derives the compact channel width $\text{c\_width}_{m_i}^c$.
The MBS algorithm takes the pre-trained model $F_n()$ with $n$ convolution layers and $N$ training images as input.
The convolution results of the pre-trained model $F_n()$ for the training images are utilized for estimating the scaling factors.
The inner loop from steps $2$ to $4$ collects non-zero statistics of the ReLU outputs $p_j$. 
The steps after the inner loop (steps $5$ and $6$) take an average over $N$ training instances, and then derive the effective flop for each convolution layer $c_j$.

The macroblock process starts from step $8$.
The total and base effective flops for each macroblock are initialized in steps $9$ and $10$, respectively.
The MBS algorithm first tallies the total and base effective flops for each macroblock (steps $11$ to $18$).  
Afterwards, MBS computes the redundant ratio $r_i$ for macroblock $m_i$ (steps $19$ to $24$).  
The scaling factor $\beta_i$ is derived from redundancy ratio $r_i$ in step $20$.
After $\beta_i$ has been computed, MBS estimates the compact channel width $\text{c\_width}_{m_i}^c$ for each macroblock $m_i$ in step $25$.

After Algorithm \ref{alg-mb-scaling} outputs the new set of channel widths, the CNN is retrained with the new channel setting to generate a more compact model $F'_n()$.
Instead of fine-tuning, the reason why we retrain the model with the new channel setting is that fine-tuning the amended model with inherited weights is no better than training it from scratch, which is consistent with the observation in \cite{rethink-pruing}.
In the experimental section, we will evaluate the effectiveness of MBS by examining the performance (prediction accuracy and model-size reduction) achieved by $F'_n()$ over $F_n()$.
 
The pre-trained model $F_n()$ has $n$ convolution layers with $n_p$ parameters and $N$ training images.
The required complexity of MBS consists of two parts: the non-zero statistics $p_j$ collection (from steps $1$ to $7$) and the redundancy $r_i$ estimation (from steps $8$ to $26$).
In the first part, we collect $p_j$ by inferencing the $N$ training images, which the statement in step $3$ can be absorbed into the forward pass of the pre-trained model. 
Hence, the computational complexity is $O(n_p \times N)$.
The second part traverses all the convolution layers of the pre-trained model for deriving the compact model.
The complexity of the second part is $O(M \times n)$ since we have already derived each $e_{c_j}$ from the first part.
The wall-clock time of the first part usually takes $50$ minutes on a PC with NVIDIA $1080$-Ti for pre-trained MobileNet on ImageNet.
Notice that we only have to conduct the first part once for each pre-trained model.
The wall-clock time of the second part is negligible, which is less than one second on the same PC.

\begin{figure*}[t]

	\begin{center}		
		\subfigure[Accuracy vs. Reduction ($\%$)]{\includegraphics[width=0.4\textwidth]{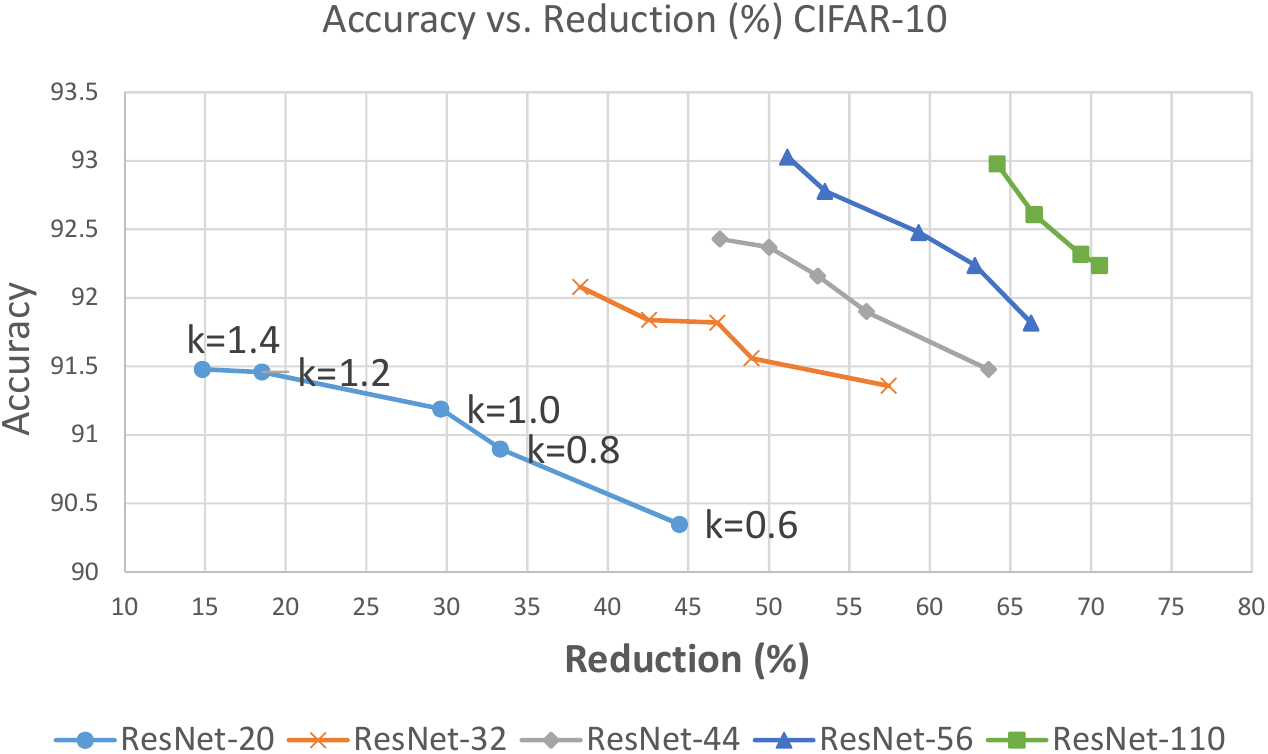}}
		\subfigure[Accuracy vs. Bitrate (KB)]{\includegraphics[width=0.4\textwidth]{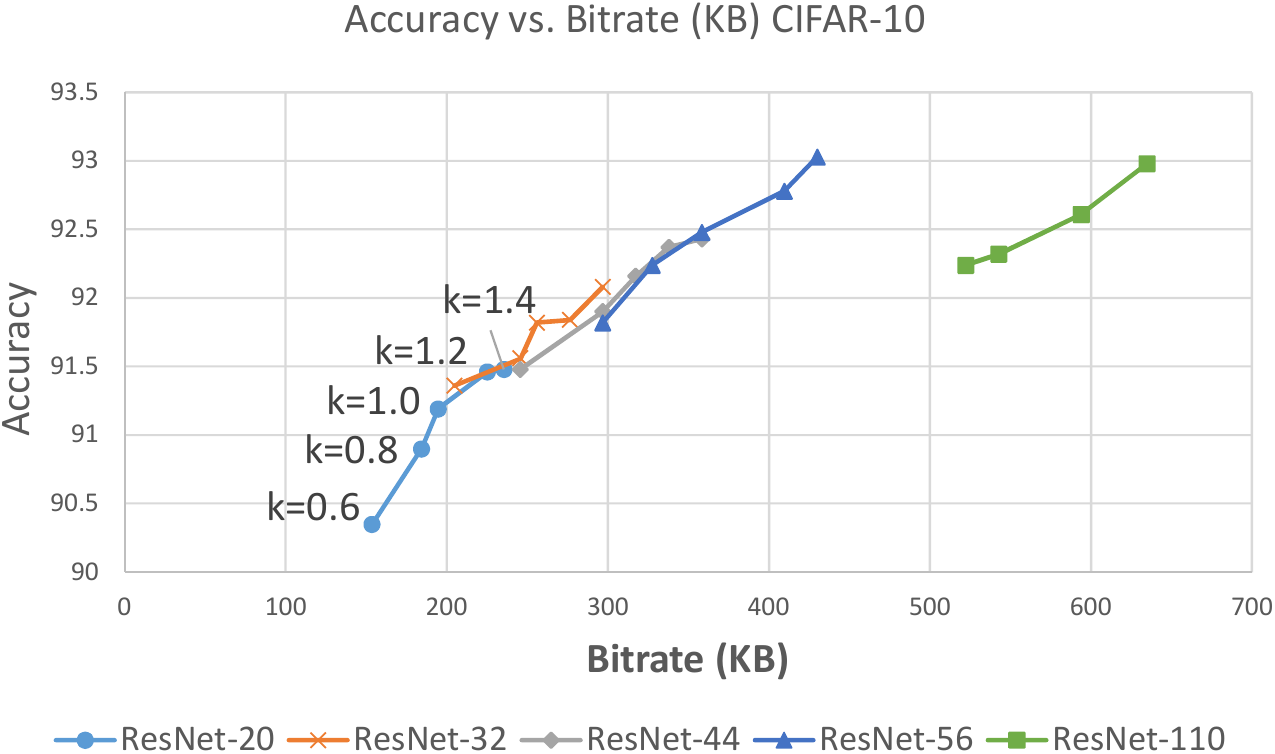}}
		
		\caption{Tradeoff between accuracy and model reduction under different receptive field settings. The value $k$ for $z=k\times L$ ranges from $1.4$ to $0.6$ with step $0.2$ for (a) accuracy vs. model reduction ratio ($\%$) and (b) accuracy vs. model size (KB).}
\label{fig-acc-field}
\end{center}
\end{figure*}

\section{Experiments}

\begin{table}[t]
	\caption{Model reduction results of ResNet on CIFAR-$10$.}
	\label{tab-cifar}
	\begin{center}
		\begin{tabular}{llc}
			\toprule
			Model & Acc. [Diff.]  & Reduction  \\
			\midrule
			\midrule
			ResNet-$20$ & $91.86\%$  & -  \\
			MBS ($L$) & $91.19\%$ [$0.67$]  & $29.63\%$   \\
			\midrule
			ResNet-$32$ & $92.24\%$ &  -  \\
			MBS ($L$) & $91.82\%$ [$0.42$] &  $46.81\%$   \\
			\midrule
			ResNet-$44$ & $92.85\%$ & -   \\
			MBS ($L$) & $92.16\%$ [$0.69$] &  $53.03\%$   \\
			\midrule
			ResNet-$56$ & $93.09\%$ &  -  \\
			MBS ($L$) & $92.48\%$ [$0.61$] &  $59.30\%$  \\
			\midrule
			ResNet-$110$ & $93.58\%$ &  -   \\
			MBS ($L$) & $92.61\%$ [$0.97$] &  $66.47\%$   \\
			\midrule
			ResNet-$1202$ & $94.04\%$ &  -  \\
			MBS ($L$) & $93.06\%$ [$0.98$] &  $72.71\%$  \\
			\midrule[2pt]
			ResNet-$110$ & $93.58\%$ &  -   \\
			\cite{prune-filter} $110$-A & $93.55\%$ [$0.03$] & $2.30\%$   \\
			\cite{prune-filter} $110$-B & $93.30\%$ [$0.28$] &  $32.40\%$   \\
			\cite{nisp} NISP & $93.35\%$ [$0.23$] &  $43.25\%$   \\
			\textbf{MBS} ($3.4 \times L$) & $93.47\%$ [$0.11$] &  $\mathbf{50.29\%}$   \\
			\bottomrule
		\end{tabular}
	\end{center}
\end{table}

We applied MBS to various CNNs on CIFAR-$10$ and ImageNet to evaluate its effectiveness in model reduction.
Our experiments aim to answer three main questions:
\begin{enumerate}
\item How aggressively can one reduce the size of a CNN model without significantly degrading prediction accuracy? (Section $4.1.1$)
\item Can MBS work effectively with deep and already highly compact CNN models? (Section $4.2.2$)
\item Can MBS outperform competing model-reduction schemes? (Sections $4.1.2$ and $4.2.1$)
\end{enumerate}

\begin{table*}[t]
	\caption{Model reduction results of CNN models with standard convolution on ImageNet.}
	\label{tab-imagenet}
	\begin{center}
		\begin{tabular}{lllccr}
			\toprule
			Model & Top-$1$ [Diff.] & Top-$5$ [Diff.] & Parameters ($\times 10^6$) & Reduction  & Configuration \\
			\midrule
			\midrule
			
			ResNet-$18$ & $69.76\%$ & $89.08\%$ & $11.69$  & - & $[64, 128, 256, 512]$ \\
			MBS ($L$) & $69.40\%$ [$0.36$] & $88.88\%$ [$0.20$] & $9.94$  & $14.97\%$  & $[64, 128, 256, 453]$ \\
			\midrule
			ResNet-$101$ & $77.37\%$ & $93.56\%$ & $44.55$  & - & $[64, 128, 256, 512]$ \\
			MBS ($L$) & $76.66\%$ [$0.72$] & $93.19\%$ [$0.37$] & $21.53$  & $51.67\%$  & $[64, 128, 174, 337]$ \\
			\midrule
			DenseNet-BC-$121$ & $74.65\%$ & $92.17\%$ & $7.98$  & - & $\beta = [1, 1, 1, 1]$ \\
			MBS ($L$) & $74.35\%$ [$0.20$] & $91.92\%$ [$0.25$] & $6.04$  & $24.31\%$  & $\beta = [1, 0.987, 0.832, 0.809]$ \\
			\bottomrule
		\end{tabular}
	\end{center}
\end{table*}

\subsection{Results on CIFAR-10 Dataset}
\label{exp-setup}

CIFAR-$10$ consists of $50$k training images and $10$k testing images of $10$ classes.
We follow the training settings in \cite{densenet}: batch size is $128$, weight decay is $10^{-4}$, and learning rate is set to $0.1$ initially and divided by $10$ at the $50\%$ and $75\%$ of the total training epochs, respectively.

\subsubsection{Accuracy and Reduction Tradeoff on ResNet}

We evaluated the effect of setting different receptive field size threshold $z$ on prediction accuracy on CIFAR-$10$ with ResNet.
The threshold $z$ is set to $z = k \times L$, which $k$ ranges from $1.4$ to $0.6$ with step size $0.2$ (from the leftmost point to the rightmost point for each line in Figure \ref{fig-acc-field}(a)).

Figure \ref{fig-acc-field}(a) shows two results. 
The $x$-axis depicts model reduction ratio from low to high, and the $y$-axis prediction accuracy. 
We first observe that on all ResNet models (ResNet-$20$, $32$, $44$, $56$, and $110$), the more number of enhancement layers (i.e., MBS employing smaller $z$ value, see the five $z$ values on each line from large on the left to small on the right), the better the model reduction ratio.  
Second, the tradeoff between model reduction and prediction accuracy exhibits in all ResNet models, as expected. 

Figure \ref{fig-acc-field}(b)
provides an application designer a handbook to guide selecting 
the receptive field setting that can fulfill the design goal.  If accuracy out-weights model size, a larger $k$ is desirable (i.e., fewer enhancement layers).  
If model size is the primary concern for power-conservation and frame-rate improvement (e.g., video analysis requiring $30$ fps), then the designer can select a smaller $k$.  
For instance, on ResNet-$32$, the bitrate of $k = 0.6$ is $200$ KB, but the bitrate of $k = 1.4$ is $300$ KB.
From the figure ResNet-$56$ seems to be a better choice than ResNet-$110$,
since given the same accuracy requirement, ResNet-$56$ always enjoys a much
lower bitrate than ResNet-$110$.


\subsubsection{MBS vs. Other Reduction Schemes}

Table \ref{tab-cifar} compares the reduction achieved by MBS and some
representative methods. The top-half of the table lists our evaluation on all ResNet models.  For instance, MBS reduces the model size of ResNet-$1202$ significantly ($72.71\%$) with negligible accuracy drop ($0.98\%$). The bottom half of the table compares MBS with recently
published methods with the best reduction ratios.  
We set MBS at the same accuracy level, MBS achieves the highest reduction ratio ($50.29\%$).

\begin{figure}[t]
	\begin{center}
		\centerline{\includegraphics[width=0.75\columnwidth]{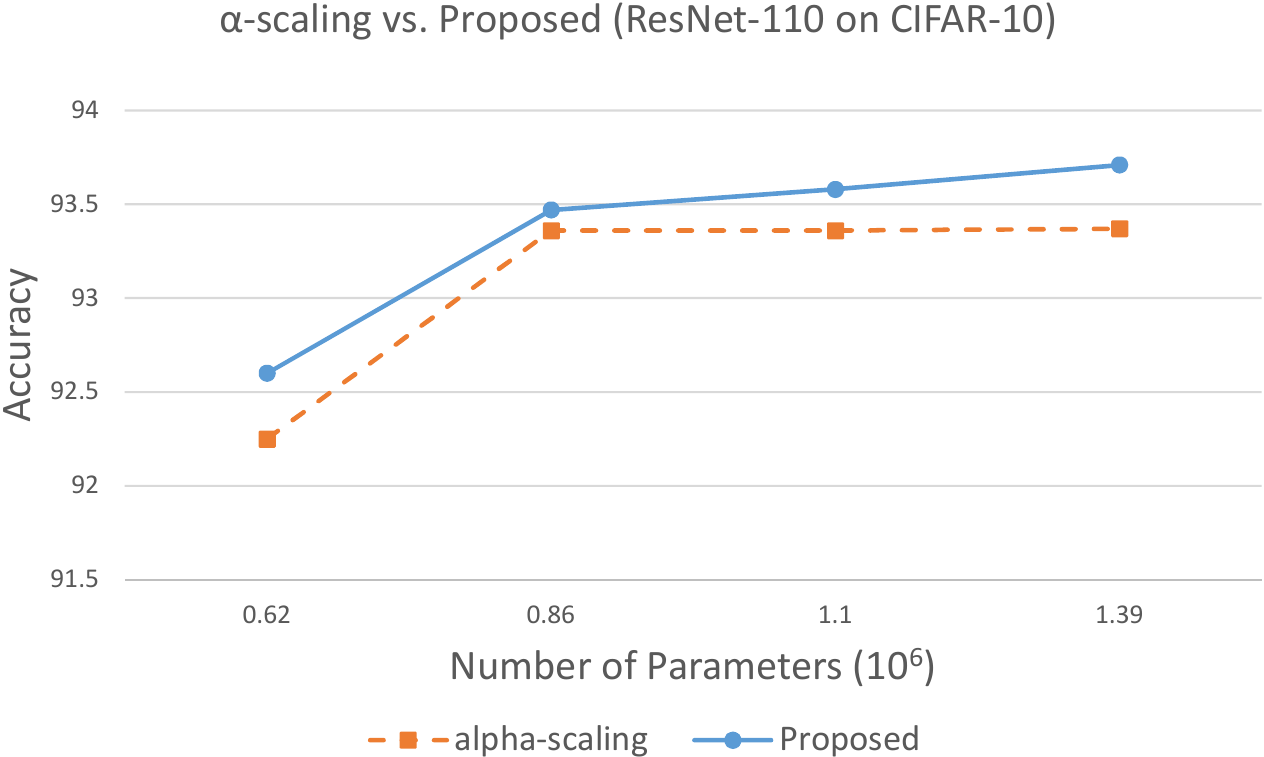}}
		\caption{The x-axis denotes the model size and the y-axis represents the prediction accuracy. The $\alpha$-scaling sets the $\alpha$ value ranges from $0.6$ to $0.9$ with step size $0.1$. The proposed algorithm compares the performance with similar model size. }
		\label{fig-alpha-compare}
	\end{center}
\end{figure}

We also compared MBS with the naive $\alpha$-scaling method 
used by ResNet.  
The $\alpha$-scaling multiplies the entire model with the same scaling factor $\alpha$, whereas MBS adaptively sets the scaling factor by the information density of each macroblock.  
Figure \ref{fig-alpha-compare} plots the range of $\alpha$ from $0.6$ to $0.9$ with step size $0.1$. 
Under the condition of having the similar resultant model size, MBS outperforms $\alpha$-scaling in prediction accuracy on four model sizes.

\begin{table*}[t]
	\caption{Model reduction results of ResNet-34 on ImageNet. }
	\label{tab-compare}
	\begin{center}
		\begin{tabular}{lllccc}
			\toprule
			{} & Top-$1$ [Diff.] & Top-$5$ [Diff.] & Parameters ($\times 10^6$) & Reduction  & Configuration \\
			\midrule
			\midrule
			ResNet-$34$ (Original) & $73.30\%$ & $91.42\%$ & $21.80$  & - & $[64, 128, 256, 512]$ \\			
			\cite{prune-filter} & $72.17\%$ [$1.13$] & - & $19.30$ & $10.80\%$  & -\\
			\cite{nisp} NISP-$34$-A & $72.95\%$ [$0.35$] & - & - & $27.14\%$  & - \\
			\cite{nisp} NISP-$34$-B & $72.31\%$ [$0.99$] & - & - & $\mathbf{43.68\%}$  & - \\
            \midrule
            MBS ($0.8\times L$) & $72.31\%$ [$0.99$] & $90.87\%$ [$0.55$] & $12.10$  & $\mathbf{44.50\%}$  & $[64, 128, 192, 359]$ \\
			\bottomrule
		\end{tabular}
	\end{center}
\end{table*}

\begin{table*}[h]
	\caption{Model reduction results of MobileNet and ShuffleNet on ImageNet.}
	\label{tab-imagenet-mobilenet}
	\begin{center}
		\begin{tabular}{lllccr}
			\toprule
			Model & Top-$1$ [Diff.] & Top-$5$ [Diff.] & Param. ($\times 10^6$) & Reduction  & Configuration \\
			\midrule
			\midrule
            ShuffleNet ($g=3$) & $65.01\%$ & $85.89\%$ & $1.88$ & -  & $[24, 240, 480, 960]$ \\
			Proposed ($L$) & $63.95\%$ [$1.06$] & $85.15\%$ [$0.74$] & $1.49$ & $\mathbf{20.74\%}$ & $[24, 240, 444, 792]$ \\
			\midrule
			MobileNet $(L=224)$ & $70.73\%$ & $89.59\%$ & $4.23$ & -  & $[32, 64, 128, 256, 512, 1024]$ \\
			Proposed ($L$) & $70.52\%$ [$0.21$] & $89.57\%$ [$0.02$] & $4.00$ & $5.43\%$ & $[32, 64, 128, 256, 512, 958]$ \\
			Proposed ($0.8 \times L$) & $69.90\%$ [$0.83$] & $89.21\%$ [$0.38$] & $3.50$ & $\mathbf{17.26\%}$ & $[32, 64, 128, 256, 474, 879]$ \\
			\midrule
			MobileNet $(L=192)$ & $68.88\%$ & $88.34\%$ & $4.23$ & -  & $[32, 64, 128, 256, 512, 1024]$ \\
			Proposed ($L$) & $68.98\%$ [$-0.10$] & $88.37\%$ [$-0.03$] & $3.93$ & $7.10\%$ & $[32, 64, 128, 256, 512, 937]$ \\
			Proposed ($0.8 \times L$) & $68.05\%$ [$0.83$] & $87.77\%$ [$0.57$] & $3.14$ & $\mathbf{25.77\%}$ & $[32, 64, 128, 256, 441, 825]$ \\
			\midrule
			MobileNetV$2$ $(L=224)$ & $71.8\%$ & $91.00\%$  & $3.47$ & - & $\beta=[1, 1, 1, 1]$ \\
			Proposed ($1.2 \times L$) & $70.81\%$ [$0.99$] & $89.89\%$ [$1.11$] & $3.14$ & $\mathbf{25.03\%}$ & $\beta=[1, 1, 0.9447, 0.7978]$ \\
			
			\bottomrule
		\end{tabular}
	\end{center}
\end{table*}

\subsection{ImageNet Dataset}

ImageNet consists of $1.28$ million training images and $50$k images for the $1,000$ class validation.  We trained all models (except for DenseNet, explained next) by $90$ epochs with batch size set as $256$. The learning rate is initially set to $0.1$ and divided by $10$ at epochs $30$ and $60$, respectively.  
For DenseNet, we trained its model by $100$ epochs with batch size $128$ and divided the learning rate by $10$ at epochs $90$ as suggested in \cite{densenet}.
The data augmentation follows the ImageNet script of PyTorch, which is the same as ResNet \cite{resnet}.  The weight decay is $10^{-4}$ for the CNNs with standard convolution (e.g., ResNet and DenseNet).
The CNNs with depth-wise separable convolution (e.g., ShuffleNet and MobileNet) set the weigh decay to $4 \times 10^{-5}$ according to the training configurations as suggested in ShuffleNet \cite{shufflenet}.

\subsubsection{Results of CNNs with Standard Convolution}
Table \ref{tab-imagenet} shows that MBS is flexible to work with different CNN designs including very
deep and complex CNN models such as ResNet-$101$ and DenseNet-$121$.
As shown in Table~\ref{tab-cifar}, 
MBS can work with different depth configurations of ResNet
on the CIFAR-10 dataset. Table \ref{tab-imagenet} further 
shows consistent results when working on ImageNet.
MBS achieves $51.67\%$ model reduction for 
ResNet-$101$, while maintaining the same prediction accuracy.
On a highly optimized deep model DenseNet-$121$ (a version of DenseNet-BC-$121$ defined in \cite{densenet}), which has bottleneck modules and transition layers already highly compressed by $50\%$.
MBS still can achieve additional $24.31\%$ model 
reduction with negligible accuracy loss. 

To exhaustively compare with all prior works, we also conducted experiments with ResNet-$34$. We divided the learning rate by $10$ at epoch $90$ and trained the reduced ResNet-$34$ with additional $10$ epochs as a simplified fine-tune process.
The state-of-the-art method NISP-$34$-B did not specify its computation complexity in \cite{nisp}. However, additional preprocessing steps are  required to derive the filter's importance.
For a model with $n_p$ parameters and $N$ training images, these preprocessing steps require $O(n_p \times N)$ to acquire the CNN outputs and sort the features with $O(N^{2.37})$ \cite{inf-fs}.
Table \ref{tab-compare} shows that MBS slightly outperforms NISP-$34$-B on ResNet-$34$ (by $0.8\%$ at the same accuracy level) and the complexity is only $O(n_p \times N)$.



\subsubsection{Results of CNNs with Depth-wise Convolution}

We applied MBS to two CNN models with depth-wise convolution structures, ShuffleNet and MobileNet.
The depth-wise convolution structure already reduces CNN model size significantly.
Table \ref{tab-imagenet-mobilenet} shows that MBS can further reduce these highly compact models.
On ShuffleNet, MBS reduces the model size by additional $20.74\%$ with negligible distortion.
The depth-wise convolution and the unique shuffling operation of ShuffleNet would increase the difficulty of the objective function formulation for the optimization-based methods. On the contrary, MBS can simply estimate the channel-scaling factor for each CNN macroblock and perform model reduction. 

We also evaluated MBS with MobileNet at different input image resolutions.
Table \ref{tab-imagenet-mobilenet} shows that MBS 
achieves $17.26\%$ and $25.77\%$ reduction on $L = 224$ and $L = 192$, respectively. 
Notice that when we set $z=L$, the prediction accuracy of MobileNet-$192$ improves slightly. This result suggests a 
possible smaller threshold value of $z$ for MobileNet.
Hence, we applied a slightly more aggressive setting of 
$z = 0.8 \times L$, which achieved a $25.77\%$ model-size 
reduction. 
Furthermore, our method is also applicable to the state-of-the-art compact CNN, MobileNetV$2$\footnote{From TensorFlow GitHub.}, and achieves $25.03\%$ reduction with negligible distortion.

\section{Conclusion}

We proposed a novel method to estimate the channel-scaling factor for each CNN macroblock. 
Our proposed MBS algorithm reduces model size guided by an information density surrogate without significantly degrading class-prediction accuracy.
MBS is flexible in that it can work with various CNN models (e.g., ResNet, DenseNet, ShuffleNet and MobileNet), and is also scalable in its ability to work with ultra deep and highly compact CNN models (e.g., ResNet-$1202$). 
MBS outperforms all recently proposed methods to reduce model size at low computation complexity. With an adjustable receptive field parameter, an application designer can determine a proper tradeoff between prediction accuracy and model size (implying DRAM size and power consumption) by looking up a tradeoff table similar to the table presented in Figure~\ref{fig-acc-field}. 

\section*{Acknowledgment}
We would like to thank Tzu-Wei Sung for his help in performing experiments on MobileNetV$2$.

{\small
\bibliographystyle{ieee}
\bibliography{receptive_field_ref}
}

\end{document}